\documentclass[times, review, 10pt]{elsarticle}

\usepackage{amssymb}
\usepackage{amsmath}
\usepackage{booktabs}
\usepackage{multirow}
\usepackage{multicol}
\usepackage{color}
\usepackage{subcaption}

\journal{Pattern Recognition}

\begin{document}

\begin{frontmatter}

\title{CLIC: Contrastive Learning Framework for Unsupervised Image Complexity Representation}

\author[1]{Shipeng Liu}
\ead{lsp@xauat.edu.cn}

\author[2,3]{Liang Zhao\corref{cor1}}
\ead{zhaoliang@xauat.edu.cn}

\author[2]{Dengfeng Chen}
\ead{chdengf@xauat.edu.cn}

\cortext[cor1]{Corresponding author}

\affiliation[1]{
            department={School of Mechanical and Electrical Engineering},
            organization={Xi'an University of Architecture and Technology},
            city={Xi'an},
            postcode={710055}, 
            state={Shaanxi},
            country={China}}
\affiliation[2]{
            department={School of Information and Control Engineering},
            organization={Xi'an University of Architecture and Technology},
            city={Xi'an},
            postcode={710055}, 
            state={Shaanxi},
            country={China}}
\affiliation[3]{
            organization={Shaanxi Key Laboratory of Geotechnical and Underground Space Engineering},
            city={Xi'an},
            postcode={710055}, 
            state={Shaanxi},
            country={China}}

\begin{abstract}
As a fundamental visual attribute, image complexity significantly influences both human perception and the performance of computer vision models. However, accurately assessing and quantifying image complexity remains a challenging task. (1) \textit{Traditional metrics such as information entropy and compression ratio often yield coarse and unreliable estimates.} (2) \textit{Data-driven methods require expensive manual annotations and are inevitably affected by human subjective biases.} To address these issues, we propose \textbf{CLIC}, an unsupervised framework based on \textbf{C}ontrastive \textbf{L}earning for learning \textbf{I}mage \textbf{C}omplexity representations. CLIC learns complexity-aware features from unlabeled data, thereby eliminating the need for costly labeling. Specifically, we design a novel positive and negative sample selection strategy to enhance the discrimination of complexity features. Additionally, we introduce a complexity-aware loss function guided by image priors to further constrain the learning process. Extensive experiments validate the effectiveness of CLIC in capturing image complexity. When fine-tuned with a small number of labeled samples from IC9600, CLIC achieves performance competitive with supervised methods. Moreover, applying CLIC to downstream tasks consistently improves performance. Notably, both the pretraining and application processes of CLIC are free from subjective bias.
\end{abstract}




\end{frontmatter}

\section{Introduction}
\label{sec:intro}
Image Complexity (IC) describes the level of visual complexity present in an image—a concept that is inherently abstract but intuitively understandable by humans \cite{1}. Subjectively, IC is reflected in the diversity of visual elements, structural intricacy, and information density within an image, all of which influence the cognitive load required for human interpretation \cite{2}. Objectively, IC can be characterized by various visual features, including color variations, texture distributions, and detail richness \cite{3}. Given its intrinsic connection to visual perception, IC has emerged as a critical attribute in numerous computer vision (CV) tasks. In recent years, the notion of \textit{computable image complexity} has been introduced, aiming to algorithmically simulate and quantify human perception of image complexity. Accurate quantification and automatic prediction of IC have shown significant potential in a range of applications, such as image understanding \cite{4}, object detection \cite{5}, UI design \cite{6}, text detection \cite{7}, and image enhancement \cite{8}. Therefore, accurate assessment and modeling of image complexity are essential not only for aligning with human perception but also for enhancing the performance of downstream CV tasks (Sec.~\ref{sec:priliminarily-2}). A reliable IC estimation facilitates more effective and perceptually aligned processing in various computer vision pipelines.

\begin{figure}[t!]
\centering
\includegraphics[width=0.65\columnwidth]{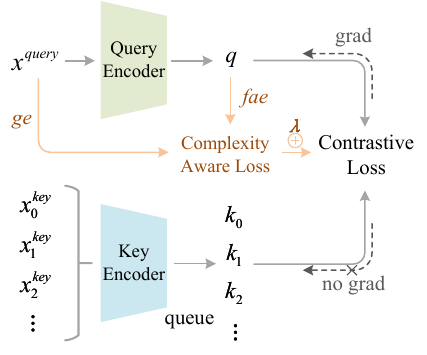}
\caption{The \textbf{CLIC framework} comprises a \textbf{query encoder} and a \textbf{key encoder} that share the same architecture. The parameters of the query encoder are updated via standard \textbf{backpropagation}, while the key encoder is updated using \textbf{momentum-based} updates. Each image in a mini-batch is pre-computed with its global entropy (\textit{ge}) as \textbf{prior information}. To guide the model towards learning intrinsic image features rather than category- or object-specific attributes, we extract the feature activation energy (\textit{fae}) from the last down-sampling layer (or the final stage of the encoder). This \textit{fae} is then combined with the image's \textit{ge} to compute the \textbf{Complexity-Aware Loss}. This loss encourages the model to focus on structural and informational content that reflects image complexity, rather than semantic content.}
\label{fig1-simple-arch}
\end{figure}

To assess or predict image complexity, early research primarily focused on heuristic-based metrics such as information entropy \cite{9}, image compression ratio \cite{10}, and classical machine learning techniques including SVMs, random forests, and BP neural networks \cite{11,12}. However, these methods largely rely on hand-crafted features, which limit their capacity to generalize image complexity (IC) assessment in real-world scenarios \cite{ic9600}. In recent years, deep convolutional neural networks (CNNs) have shown strong representational power and generalization ability, enabling more effective modeling of human subjective perception in a data-driven manner—particularly in related tasks such as image quality assessment \cite{14}. This progress has led to the development of large-scale IC evaluation datasets (e.g., IC9600 \cite{ic9600}) and corresponding deep learning-based IC prediction models (e.g., ComplexityNet \cite{15}, ICNet \cite{ic9600}, ICCORN \cite{16}, etc.). Nevertheless, constructing such large-scale supervised datasets remains labor-intensive, as IC scores must be manually annotated for each image, posing a major bottleneck in the scalability and practicality of supervised IC learning approaches.

To mitigate the high cost and inherent subjectivity associated with manual annotations, a natural alternative inspired by unsupervised learning is to leverage unlabeled datasets for image complexity (IC) representation learning. Once meaningful representations are learned, IC scores can be obtained by fine-tuning the model on a small set of labeled samples. In the realm of unsupervised learning, contrastive learning methods—such as MoCo \cite{moco}—employ momentum encoders and dynamic dictionaries to address the challenge of insufficient negative samples. These methods aim to maximize the similarity between positive pairs while minimizing the similarity between positive and negative pairs. The success of MoCo has catalyzed the development of numerous enhanced approaches, including MoCo v2 \cite{19}, SimCLR \cite{20}, BYOL \cite{21}, and SwAV \cite{22}. These methods have significantly advanced the field of self-supervised contrastive learning and demonstrated strong performance across a wide range of computer vision tasks.

While contrastive learning has shown great success in unsupervised representation learning, directly applying it to image complexity (IC) representation poses significant challenges. (1) \textit{IC features are abstract and distinct from general image features}; they should remain invariant across different image categories and not be dominated by category-specific attributes. (2) \textit{Defining positive and negative pairs for IC representation is non-trivial}, as conventional contrastive learning on classification datasets (e.g., ImageNet) tends to induce category bias into the learned representations. Motivated by these challenges, we propose the \textbf{CLIC} framework (Fig.~\ref{fig1-simple-arch}), which leverages contrastive learning to effectively learn image complexity representations in an unsupervised manner. Specifically, CLIC employs a dual-encoder structure consisting of a \textbf{query encoder} and a \textbf{key encoder}, updated via gradient descent and momentum, respectively. To address the difficulty of sample pair construction in the context of IC, we design a novel positive and negative sample selection strategy tailored for complexity representation learning. Moreover, we incorporate image prior information to formulate a \textbf{Complexity-Aware Loss}, which explicitly suppresses category-related interference and guides the model to focus on structural and informational cues associated with complexity. Extensive experiments validate the effectiveness of the proposed framework. CLIC, when fine-tuned on a limited number of labeled samples from IC9600 \cite{ic9600}, achieves performance competitive with fully supervised methods. Furthermore, applying CLIC to various downstream tasks yields significant performance improvements, highlighting its practical value and broad applicability in real-world scenarios.

In summary, our main contributions are as follows:
\begin{itemize} 
\item \textbf{CLIC}, a novel contrastive learning framework for unsupervised image complexity representation, is proposed. This framework enables the learning of complexity-related features without relying on manual scoring labels.
\item A tailored positive and negative sample selection strategy is designed, specifically for image complexity. The augmented view that most closely preserves the original image's complexity is treated as the positive sample for contrastive loss calculation.
\item A Complexity-Aware Loss, based on global entropy as prior knowledge, is introduced. This loss function optimizes the model’s ability to capture complexity by penalizing discrepancies between the learned representation and the image’s inherent complexity.
\item Extensive experiments demonstrate the effectiveness of CLIC. The simple weighted application of CLIC encoder features significantly improves performance across various downstream computer vision tasks.
\end{itemize}

\section{Related Work}
\label{sec:related_work}

\textbf{Image Complexity Estimation.}
Modeling image complexity has long been a topic of interest in both computer vision and human perception. Early works relied on handcrafted features, such as entropy~\cite{28}, symmetry~\cite{23,34}, spatial layout~\cite{25}, and compressibility~\cite{29,30}, to quantify complexity in a rule-based manner. These methods provide interpretable signals but struggle to generalize to diverse real-world images.

With the advent of deep learning, learning-based approaches have emerged to model visual complexity. Nagle et al.~\cite{32} and Saree et al.~\cite{33} utilized CNN features to regress complexity scores obtained from human annotations. ComplexityNet~\cite{34} leveraged local symmetry descriptors, while Feng et al.~\cite{ic9600} proposed ICNet, a weakly supervised model trained on a curated complexity dataset (IC9600). Extensions of ICNet have incorporated ordinal regression~\cite{16} or segmentation statistics~\cite{35,36,37} to better capture structural aspects. Despite their effectiveness, these models require either human supervision or proxy labels, limiting their scalability and adaptability to new tasks.

\textbf{Contrastive Representation Learning.}
Contrastive learning has become a standard paradigm for self-supervised visual representation learning. Methods such as InstDisc~\cite{38}, MoCo~\cite{moco}, and SimCLR~\cite{20} learn representations by pulling together augmented views of the same instance while pushing apart different instances. Further refinements like BYOL~\cite{21}, SimSiam~\cite{40}, and SwAV~\cite{22} improve training stability and efficiency by removing the need for negative samples or incorporating clustering. These methods have demonstrated strong performance in downstream recognition tasks.

However, contrastive learning has primarily been explored in the context of semantic representation learning, where the objective is to encode discriminative features for downstream classification or detection. Existing frameworks are not designed to capture structural complexity, which is often orthogonal to semantic content.

\textbf{Our Perspective.}
In this work, we explore contrastive learning from a new perspective—learning to represent visual complexity in an unsupervised manner. To the best of our knowledge, this is the first attempt to adapt contrastive learning for image complexity modeling. Our proposed CLIC framework introduces a complexity-aware sampling strategy and a complexity-aligned loss function, guiding the encoder to capture structural and perceptual difficulty rather than semantic categories. This enables learning complexity representations without reliance on labeled data, offering a scalable and task-agnostic solution for perceptual modeling.

\section{Preliminary}
\subsection{Definition of Image Complexity}
Image complexity is inherently difficult to define. Traditional manual methods, such as image entropy and image compression ratio, offer some metrics, but these are primarily heuristic and often inaccurate for assessing complexity, with limited generalization ability. Machine learning methods typically rely on supervised learning to model image complexity, which requires costly annotations and inevitably incorporates human subjective biases. Regardless of whether manual or machine learning methods are employed, image complexity modeling can be formally expressed as in Eq.~(\ref{eq1}), where ${x_i}$ represents an image from a dataset and $f( \cdot )$ denotes the mapping from this image to its corresponding image complexity $IC_i$, with $IC_i$ taking values within the range of 0 to 1.

\begin{equation}
    I{C_i} = f({x_i})
    \label{eq1}
\end{equation}

\begin{figure}[t!]
	\centering
	\begin{minipage}{0.49\columnwidth}
		\centering
		\includegraphics[width=0.9\columnwidth]{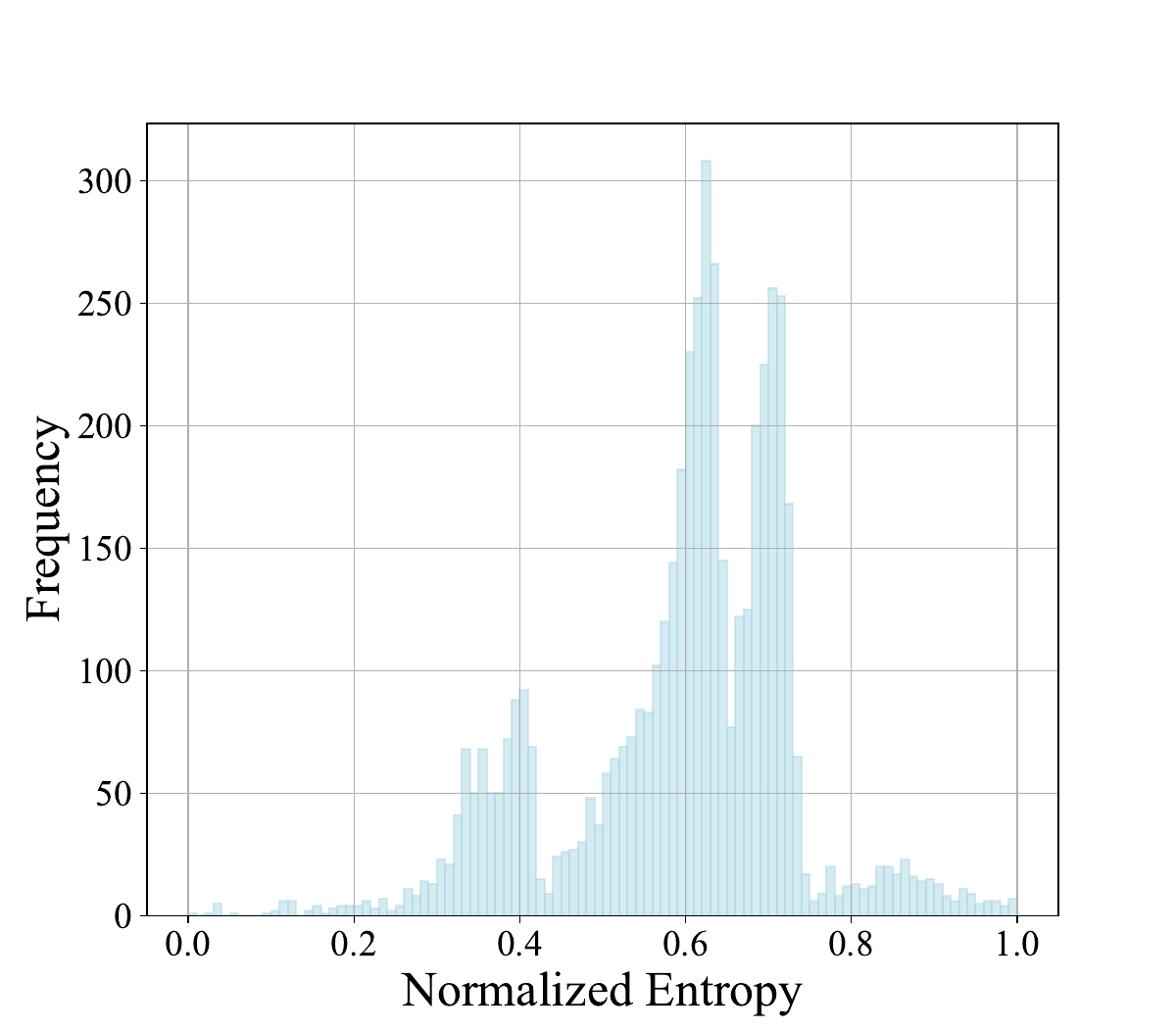}
	\end{minipage}
	\begin{minipage}{0.49\columnwidth}
		\centering
		\includegraphics[width=0.9\columnwidth]{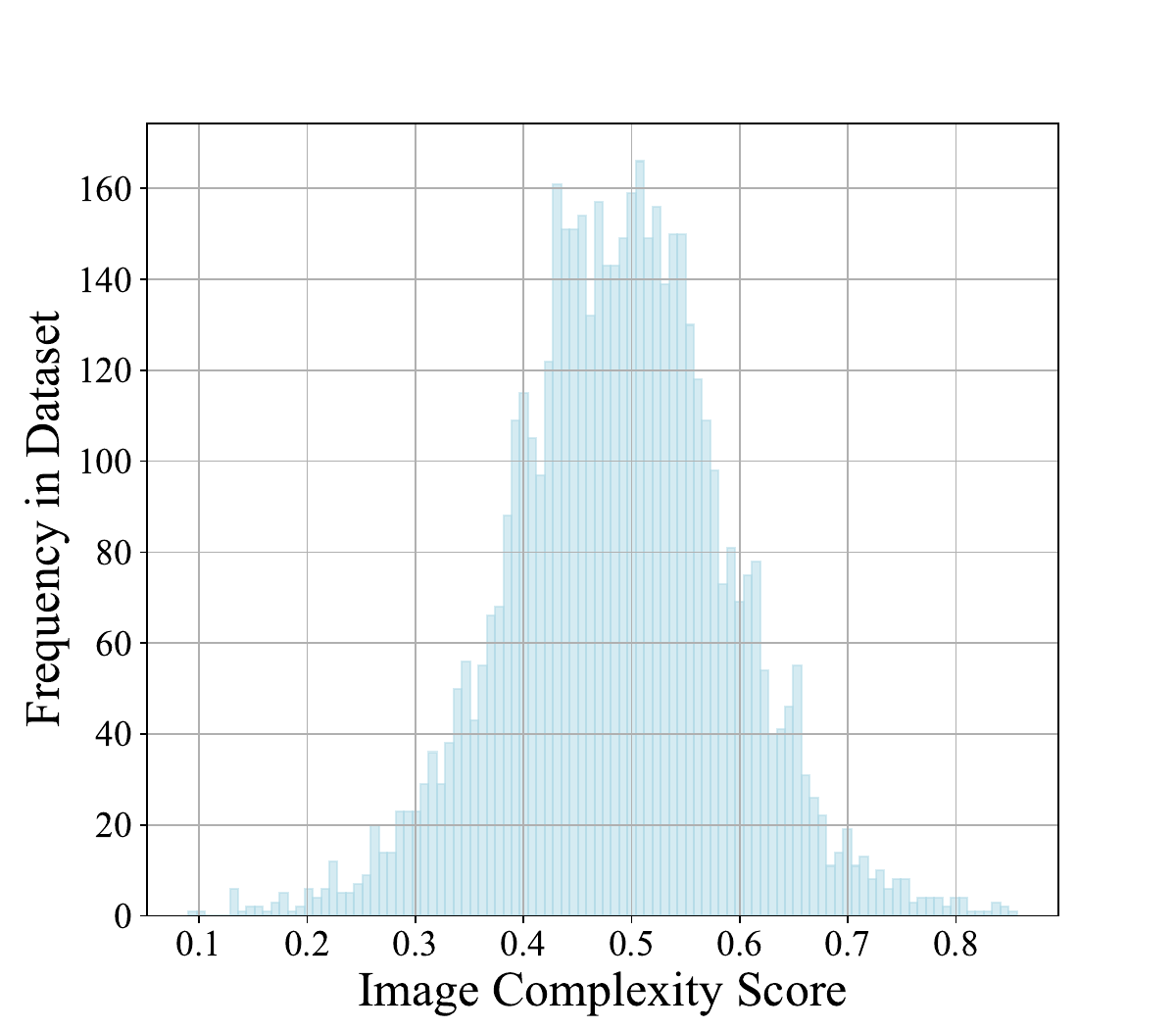}
	\end{minipage}
	\begin{minipage}{0.49\columnwidth}
		\centering
		\includegraphics[width=0.9\columnwidth]{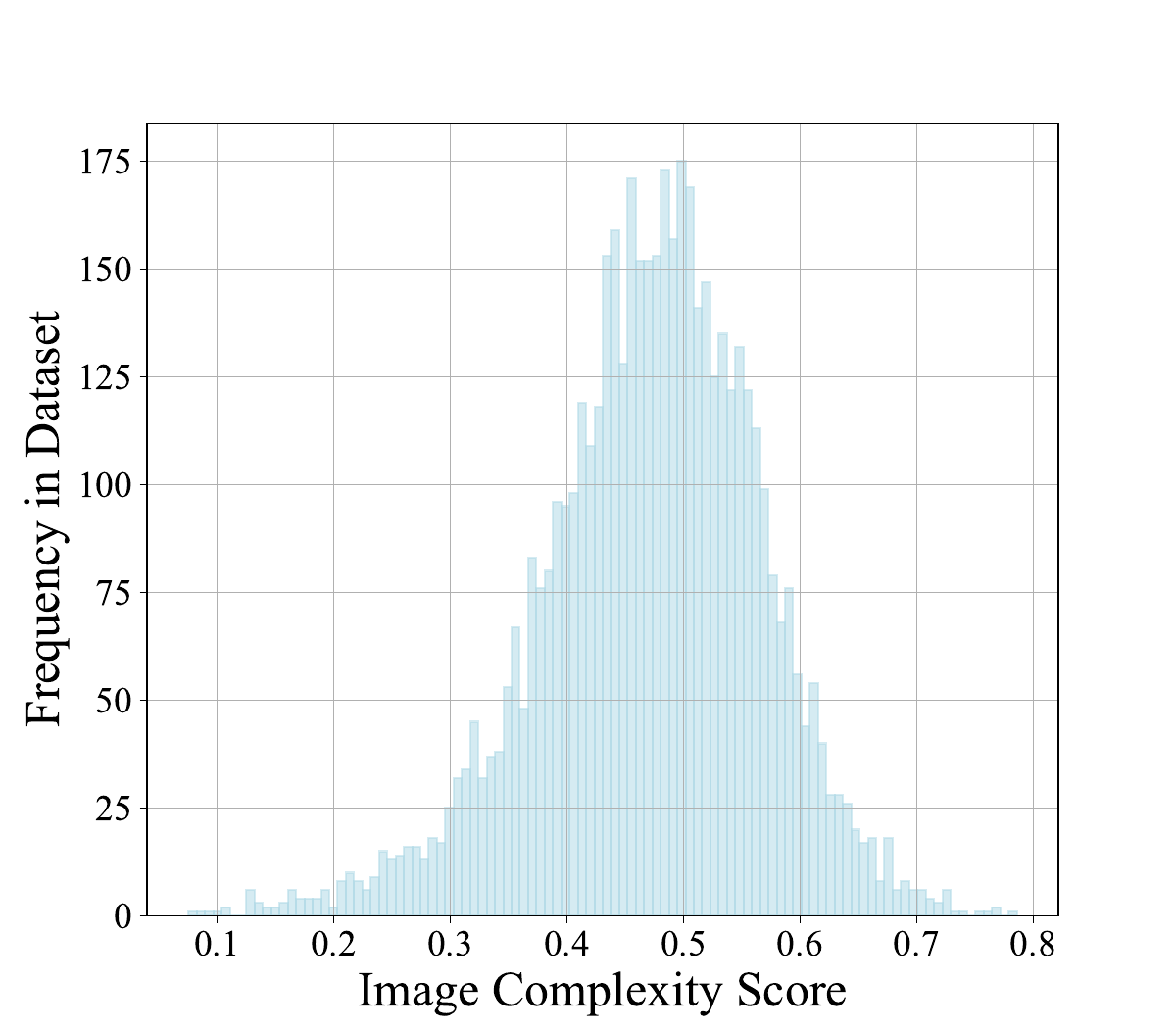}
	\end{minipage}
	\begin{minipage}{0.49\columnwidth}
		\centering
		\includegraphics[width=0.9\columnwidth]{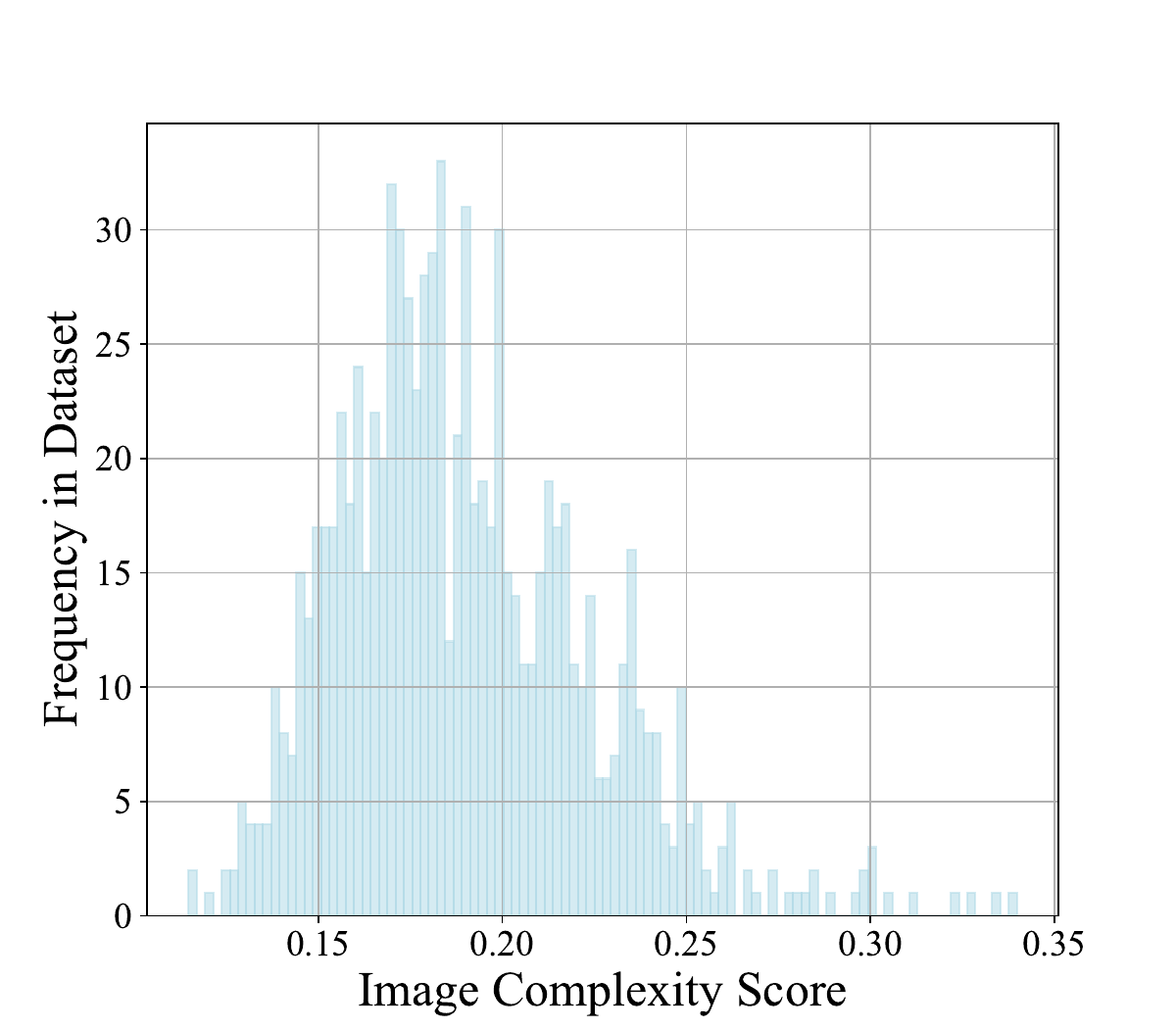}
	\end{minipage}
\caption{\textbf{Image Complexity Distribution of Datasets.} \textbf{(a)} global entropy of MS COCO \cite{mscoco}. \textbf{(b)} ICD of MS COCO \cite{mscoco}. \textbf{(c)} ICD of PASCAL VOC \cite{46}. \textbf{(d)} ICD of Cornell Grasp \cite{44}. The global entropy is obtained by Eq.(2). ICD stands for image complexity distribution, which represents the true complexity of each image. In this work, it is obtained by conducting statistics on the inference of each image by a well-trained ICNet.}
\label{fig2-icd}
\end{figure}

After analyzing the image complexity across several datasets, it is observed that, in general, the Image Complexity Distribution (ICD) of a dataset approximates a normal distribution, i.e., $IC_i \sim N(\mu, \sigma^2)$, suggesting that most images tend to exhibit a medium level of complexity. As shown in Fig.~\ref{fig2-icd}, the ICD of the dataset is computed using both image entropy and machine learning methods, and the resulting distribution curve closely follows the normal distribution pattern. Notably, datasets with a normal characteristic ICD are predominantly composed of natural scene images, such as street scenes, indoor/outdoor environments, and fields. However, certain datasets, like the Cornell Grasp Dataset \cite{44}, exhibit a high concentration of ICD values around 0.17. This characteristic is one of the reasons why tasks involving such datasets, particularly the 2D target detection task on the Cornell Grasp Dataset, tend to be easier to learn.

\begin{table}[t!]
    \centering
    \setlength{\tabcolsep}{1mm}{
    \begin{tabular}{c|lllc}
    IC Range          & AP         & AP$_{50}$       & AP$_{75}$       &  \\ \specialrule{1.2pt}{0pt}{0pt} 
    All               & 40.5       & 59.3       & 43.8       &  \\
    \textless{}0.3    & 41.2$_{(+0.7)}$ & 60.3$_{(+1.0)}$ & 44.6$_{(+0.8)}$ &  \\
    \textgreater{}0.7 & 39.6$_{(-0.9)}$ & 58.0$_{(-1.3)}$ & 42.6$_{(-1.2)}$ & 
    \end{tabular}}
    \caption{\textbf{Results of YOLOX-S on MS COCO with different IC range.} The training details for YOLOX-S were maintained consistent with the official implementation. All represents the original MS COCO val set, with the IC range from 0 to 1.}
    \label{tab1}
\end{table}

\subsection{Impaction of Image Complexity}
\label{sec:priliminarily-2}
We utilize ICNet \cite{ic9600} for inference on the MS COCO \cite{mscoco} dataset, sampling images within the IC ranges of $\textless 0.3$ and $\textgreater 0.7$ as validation sets. Each complexity range consists of 5k images, forming two training-validation set pairs. These pairs are trained based on the pre-trained YOLOX-S \cite{47}, and the results are presented in Tab.~\ref{tab1}. The results are consistent: the average precision (AP) increases for IC $\textless 0.3$ and decreases for IC $\textgreater 0.7$, suggesting that lower image complexity leads to simpler tasks and, consequently, improved performance. This underscores the importance of accurately modeling image complexity and applying it judiciously to fundamental computer vision tasks (e.g., detection and segmentation). This phenomenon has also been demonstrated through extensive experiments by Feng et al. \cite{ic9600}.

\section{Method}
\label{sec:Method}

\subsection{Positive and Negative Samples Selection}
\label{subsec: 3.1}

\begin{figure*}[t]
    \centering
    \includegraphics[width=0.95\textwidth]{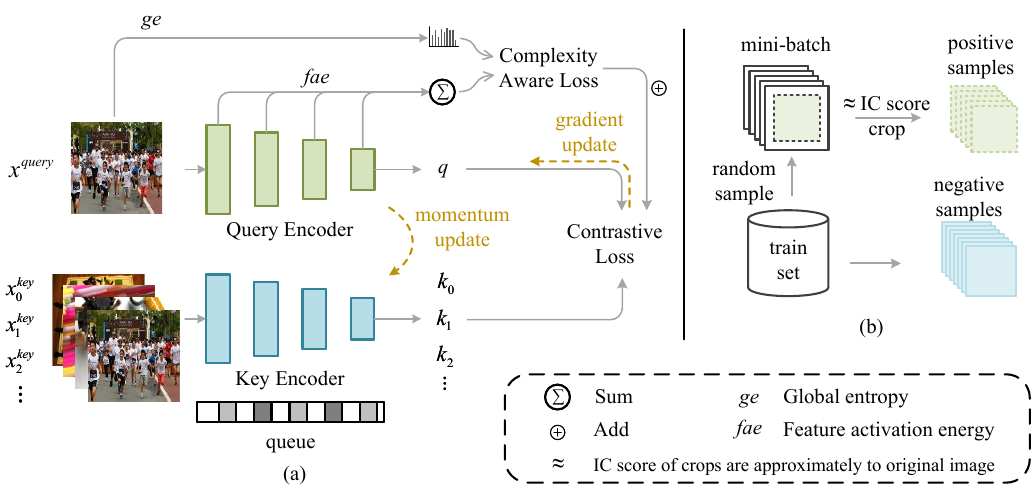}
    \caption{\textbf{Our overall architecture.} \textbf{(a) Detail structure of CLIC}. \textbf{(b) Positive and Negative Samples Selection.} We crop the mini-batch image to produce views whose image complexity is close ($\sim$ IC score) to that of the original as positive samples. Outside the mini-batch are negative samples.}
    \label{fig3-overall}
\end{figure*}

A careful selection of positive and negative samples in representation learning plays a critical role in helping the model learn meaningful image representations. This meaning is typically defined for a specific task, such as the similarity of positive sample pairs in image classification. In the context of image complexity (IC) representation learning, we propose a strategy for selecting positive and negative samples that ensures the model focuses on the complexity features of images, rather than on \textbf{content, categories, or object properties.} Unlike traditional representation learning, the selection of positive and negative samples for IC representation must be designed around the intrinsic complexity properties of the images themselves, rather than relying on category or content similarity. In fact, it is crucial to \textbf{\textit{minimize interference from such extraneous factors}} as much as possible.

In the image complexity assessment task, each image is assigned an IC score to represent its complexity. Theoretically, the IC scores of any two images are unlikely to be identical. As a result, we treat each image as belonging to a distinct class, meaning that all images except for the image itself are considered negative samples. Some methods, such as SimCLR \cite{20}, generate different views of the original image as positive sample pairs and treat images outside the mini-batch as negative samples. We adopt a similar approach (Fig.~\ref{fig3-overall}(b)), with a key distinction being that we apply special treatment to the views in the context of IC representation learning.

\begin{figure}[t]
    \centering
    \includegraphics[width=0.65\columnwidth]{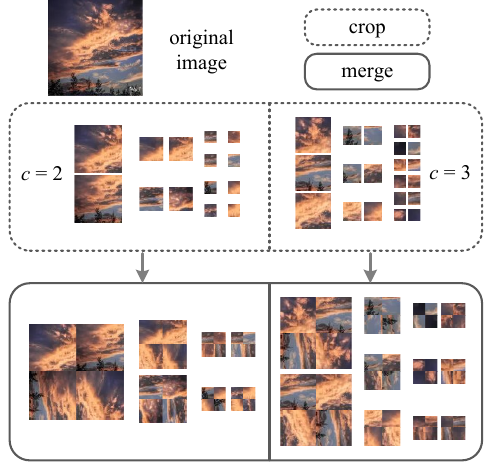}
    \caption{\textbf{Small-scale crop and merge.} Given an image of size $(h,w)$, the crop size is $(h,w)/c$, $(h,w)/2c$, and $(h,w)/4c$, the corresponding number of crops for each image is $c$, $2c$, and $4c$, respectively. Then, we randomly merge two crops of the same size and their two transformed crops.}
    \label{fig4-sscm}
\end{figure}

In the image classification task, positive sample pairs belong to the same class, meaning that sub-images obtained through random cropping from the original image are considered part of the same class as the original image. Additionally, the IC scores of the original image and the resulting view should be as similar as possible in the IC representation. To achieve this, we design cropping schemes of varying sizes (with data augmentation) and evaluate them using the Pearson correlation coefficient (PCC) \cite{49} between the views and the original image. The crops with the highest PCC are selected as the views and are defined as a positive sample pair. It is important to note that this IC score proximity refers to the fact that, from a statistical perspective, we have designed specific combinations to obtain detailed cropping schemes. As a result, cropping is directly applied during training, eliminating the need for additional computational time overhead.

Our framework (Fig.\ref{fig3-overall}(a)) does not introduce any additional definition for negative samples; instead, images outside the mini-batch are treated as negative samples, with a dynamic queue used to store negative sample features \cite{moco}. Furthermore, we process the original dataset by applying small-scale cropping and merging it in a way that introduces local image features into the model learning phase. This process enhances the learning of image complexity representations \cite{ic9600}. It should be noted that the small-scale crop and merge (Fig.\ref{fig4-sscm}) approach does not simply augment the original data volume but focuses on enhancing local features, allowing the model to learn a more localized complexity representation of the image.

\subsection{Complexity-Aware Loss with Global Entropy Prior}
Contrastive loss, commonly used in representation learning, is designed to distinguish similarities or differences between positive and negative samples. However, similarity computation in such methods may inadvertently introduce image category attributes into the model's learning process, leading to potential errors in complexity perception. Therefore, image complexity representation learning must guide the model to focus more on, or even prioritize, the fundamental attributes of an image—attributes that should be independent of category, content, or object—when extracting features. We observed an important phenomenon using some simple metrics for image complexity assessment (Tab.~\ref{tab2}). Specifically, we randomly sampled 1000 images from the ImageNet training set, computed the IC evaluation results for each of these simple metrics, and used the inference output of ICNet as the ground truth. The results showed that edge density, UAE \cite{33}, and global entropy all exhibit relatively high Pearson correlation coefficients with the ground truth (details in \ref{supp-metrics}). This finding is consistent with prior work, where scholars have demonstrated that metrics like global entropy can effectively evaluate image complexity \cite{30}.

\begin{table}[hbt!]
\centering
\setlength{\tabcolsep}{1mm}{
\begin{tabular}{c|ccccc}
PCC & UAE & ED & CR & GE & GT \\ \specialrule{1.2pt}{0pt}{0pt}
UAE & - & 0.2912 & -0.3594 & \textbf{\textcolor{magenta}{0.4971}} & \textbf{\textcolor{red}{0.5582}} \\
ED & - & - & -0.2472 & 0.3930 & \textbf{\textcolor{magenta}{0.5034}} \\
CR & - & - & - & -0.8401 & -0.3842 \\
GE & - & - & - & - & \textbf{\textcolor{magenta}{0.5259}} \\
GT & - & - & - & - & -
\end{tabular}}
\caption{\textbf{PCC of image metrics.} \textbf{UAE} is unsupervised activation energy. \textbf{ED} is edges density. \textbf{CR} is compress ratio. \textbf{GT} is ground truth.}
\label{tab2}
\end{table}

In addition, Saraee et al. \cite{33} explored the feature extraction properties of intermediate convolutional layers in deep neural networks, which are capable of capturing both low-level features (e.g., edges, texture) and high-level features (e.g., objects and scene content) of an image. They quantized several convolutional feature maps to create a mapping from feature activation energy to image complexity scores in an unsupervised manner. Interestingly, our experiments revealed that the highest Pearson correlation coefficient (PCC) was found between UAE and global entropy, compared to other combinations, and that there was a certain degree of positive correlation with the ground truth. This suggests that a relationship between global entropy and feature activation energy can be established to guide the model in learning a representation that aligns closely with the ground truth.

Motivated by these findings, we propose using an image's global entropy \textbf{\textit{ge}} as \textbf{prior information} to construct an auxiliary task for image complexity representation learning. This leads to the formulation of the Image Complexity-Aware Loss (CAL). The objective of CAL is to optimize the complexity characteristics of both positive and negative samples. The image complexity characterization task should direct the model's focus toward the image complexity features, utilizing the image's prior information to generate representations that minimize errors caused by content, category, or object attributes—errors that could hinder the model's complexity awareness. For each image in the mini-batch, its global entropy is pre-computed using the following formula:

\begin{equation}
    ge =  - \sum\limits_{i = 0}^{L - 1} {p(i) \cdot lo{g_2}(p(i))}
    \label{eq2}
\end{equation}
where $p(i)$ denotes the probability of gray value $i$, which is obtained by counting the number of pixels in each gray level $L$ (usually 256 for 8-bit gray image) and then normalized, for feature activation energy, we extract the feature maps of each downsampling (every stage in ResNet \cite{resnet} or SWin \cite{swin} series) of the query encoder to calculate the feature activation energy \textbf{\textit{fae}} and sum it:
\begin{equation}
    fae = \sum\limits_s {(\frac{1}{{h \times w \times d}}\sum\limits_{i,j,k}^{h \times w \times d} {F[i,j,k]} )}
    \label{eq3}
\end{equation}
where $F$ is the feature map, $h$, $w$, and $d$ are the feature map's height, width, and depth (number of channels), respectively. $s$ denotes the number of down-sampling. After that, MSE loss is used to construct the image complexity-aware loss $L_{CAL}$ form as follows:
\begin{equation}
    {L_{CAL}} = \frac{1}{N}\sum\nolimits_i^N {{{(x_i^{fae} - x_i^{ge})}^2}}
    \label{eq4}
\end{equation}
where $N$ is the batch size, $i \in [0, N]$. $x_i^{fae}$ and $x_i^{ge}$ denote feature activation energy \textit{\textbf{fae}} and global entropy \textit{\textbf{ge}}, respectively. 

In addition, following some representation learning methods \cite{moco,19,20}, this paper also considers InfoNCE \cite{56} loss form ${L_{IN}}$ (Eq.(\ref{eq5})) to measure positive and negative sample similarity in terms of dot product while building dictionary look-up problem for contrastive learning.

\begin{equation}
    {L_{IN}} =  - \log \frac{{\exp (q \cdot {k^ + }/\tau )}}{{\exp (q \cdot {k^ + }/\tau ) + \sum\nolimits_{{k^ - }} {\exp (q \cdot {k^ - }/\tau )} }}
    \label{eq5}
\end{equation}
where $\tau$ is temperature, $q$ is the encoded query. $k^+$ is the positive sample key and $k^-$ is the negative one. Finally, the loss of CLIC is obtained by combining $L_{CAL}$ as a regular term with InfoNCE:
\begin{equation}
    L = {L_{IN}} + \lambda {L_{CAL}}
    \label{eq6}
\end{equation}
where $\lambda$ is a hyper-parameter denoting the weight coefficient of complexity-perceived loss.

\subsection{Contrastive Learning Framework for IC Representation}
We have developed a framework for image complexity representation based on contrastive learning to efficiently learn image complexity features. The framework employs two independent encoders: the query encoder and the key encoder. The query encoder is tasked with extracting complex features from the input images, while the key encoder processes the inputs and generates a 128-dimensional vector, which is then used for contrastive learning with the feature output from the query encoder. In line with \cite{moco}, both encoders share the same network architecture, with the query encoder being updated via gradient-based optimization, and the key encoder employing momentum-based updates.
\begin{equation}
    {\theta _k} \leftarrow m{\theta _k} + (1 - m){\theta _q}
    \label{eq7}
\end{equation}
where $\theta _k$ is the parameter of the key encoder, $\theta _q$ is the parameter of the query encoder, and $m \in [0,1)$ is the momentum factor (default 0.999). 

The purpose of momentum updating is to help the model maintain a stable perception of image complexity during training by preserving the key encoder's \textit{\textbf{historical information}}, thereby mitigating the risk of losing important details due to fluctuations in the training process. During training, the query and key encoders receive different transformed views as input. Specifically, the query encoder processes views with minimal data augmentation, while the key encoder is provided with inputs that include additional data augmentation.

Furthermore, contrastive learning methods typically require large datasets to train the model and obtain the desired representations. In this work, we sample from several publicly available datasets \cite{flickr, imagenet}, using the global entropy of the images—which exhibits the highest correlation with the ground truth—as the basis for sampling. The sampling results indicate that the global entropy of the images in the training set follows a uniform distribution ($ge \in (0,1)$), suggesting that the true image complexity distribution in the training set is also close to a uniform distribution. As a result, the model is exposed to a balanced representation of image complexities of varying difficulty levels, facilitating the learning of diverse image complexity representations.

\section{Experiments}
\label{sec:Experiments}

\subsection{Setup}
\subsubsection{Training Data}
\textbf{Data Collection.} The training images are randomly sampled from Flickr-5B \cite{flickr} and ImageNet-1M \cite{imagenet}, resulting in a training set of approximately 1 million images. The image entropy of these samples follows a uniform distribution.

\subsubsection{Implementation Details} 
\textbf{Training.} Both the query encoder and the key encoder are pre-trained on ImageNet. We use a batch size of 128 across 4 NVIDIA Geforce 3090 GPUs. The initial learning rate is set to 0.03. Training proceeds for 200 epochs, with the learning rate being reduced by a factor of 10 at the 120th and 160th epochs. The momentum for updating the key encoder is set to 0.999. The softmax temperature is set to 0.07. Stochastic Gradient Descent (SGD) is used for optimization, with momentum set to 0.9 and weight decay set to 0.0001.

\textbf{Fine-tuning.} Fine-tuning of the CLIC query encoder is performed on IC9600 using ICNet \cite{ic9600}. The fine-tuning pipeline is shown in Fig.\ref{fig13} in Appendix C. During fine-tuning, the parameters of the CLIC query encoder are frozen. The batch size is 128 across 4 GPUs. SGD is used for optimization, with momentum set to 0.9, weight decay set to 0.0001, and the learning rate fixed at 0.001. All other settings are consistent with those used in ICNet.

\subsection{Comparison with Previous Results}
We compare CLIC with other methods for evaluating image complexity, both unsupervised and supervised, using the same evaluation metrics as in \cite{ic9600}, namely the Pearson Correlation Coefficient (PCC) and Spearman's Rank Correlation Coefficient (SRCC) \cite{50}. The trained CLIC query encoder serves as the backbone of ICNet, with its parameters frozen before fine-tuning on the IC9600 training set (see Tab.\ref{tab3}). It is important to note that while ICNet uses a two-branch structure, we only utilize the CLIC query encoder as a single branch.

\begin{table}[hbt!]
\centering
\begin{tabular}{cc|cc}
\multicolumn{2}{c|}{Method} & PCC↑ & SRCC↑ \\ \specialrule{1.2pt}{0pt}{0pt}
\multirow{5}{*}{unsup.} & SE \cite{31} & 0.534 & 0.498 \\
 & NR \cite{31} & 0.556 & 0.541 \\
 & ED \cite{12} & 0.569 & 0.491 \\
 & AR \cite{gu2017no} & 0.571 & 0.481 \\
 & UAE \cite{33} & \textbf{0.651} & \textbf{0.635} \\ \hline
\multirow{6}{*}{super.} & SAE \cite{33} & 0.865 & 0.86 \\
 & ComplexityNet$^\dagger$ \cite{15} & 0.873 & 0.870 \\
 & HyperIQA \cite{14} & 0.935 & 0.935 \\
 & P2P-FM \cite{ying2020patches} & 0.940 & 0.936 \\
 & ICNet$^\dagger$ \cite{ic9600} & 0.947 & 0.944 \\
 & ICCORN \cite{16} & \textbf{0.954} & \textbf{0.951} \\ \hline
\multirow{4}{*}{CLIC} & Res18 \cite{resnet} & 0.890 & 0.883 \\
 & Res50 \cite{resnet} & 0.893 & 0.887 \\
 & Res152 \cite{resnet} & 0.899 & 0.893 \\
 & Swin-B \cite{swin} & \textbf{0.913} & \textbf{0.908}
\end{tabular}
\caption{\textbf{Comparison with previous results.} $\dagger$ denotes our reproduction results, and unlabeled results are obtained from the literature \cite{ic9600}. \textbf{unsup.} denotes unsupervised methods. \textbf{super.} denotes supervised methods.}
\label{tab3}
\end{table}

The results indicate that the unsupervised methods achieve the best results in UAE, but there is a significant performance gap when compared to the supervised method, ICCORN \cite{16}. It is important to note that the unsupervised group primarily relies on traditional hand-designed heuristics. In contrast, most methods in the supervised group are based on deep learning, which significantly enhances feature extraction and leads to much better performance. While CLIC does not outperform the supervised methods, its goal is to address the high cost of manual labeling and eliminate subjective bias. When replacing the model with a more complex one, both the PCC and SRCC improve to 0.913 and 0.908, respectively, which are very close to those of the supervised methods.

\subsection{Ablation of CLIC}
To verify the validity of the individual components of CLIC, we conducted ablation studies. Each case (from a to e, as shown in Tab.\ref{tab4}) begins with an ImageNet pre-trained model. After training, the CLIC query encoder is fine-tuned on the IC9600 dataset. The results presented in the table validate the effectiveness of each individual component and demonstrate that there is no conflict between them.

\begin{table}[t]
\centering
\setlength{\tabcolsep}{1mm}{
\begin{tabular}{c|ccc|ll}
\multirow{2}{*}{Case} & \multicolumn{3}{c|}{Pretraining} & \multicolumn{2}{c}{IC9600} \\
 & pos & scm & CAL & \multicolumn{1}{c}{PCC↑} & \multicolumn{1}{c}{SRCC↑} \\ \specialrule{1.2pt}{0pt}{0pt}
(a) & \textbf{} & \textbf{} & \textbf{} & 0.759 & 0.748 \\
(b) & \textbf{\checkmark} & \textbf{} & \textbf{} & 0.796\textcolor{green}{$_{(+.037)}$} & 0.781 \\
(c) & \textbf{} & \textbf{\checkmark} & \textbf{} & 0.804\textcolor{green}{$_{(+.045)}$} & 0.788 \\
(d) & \textbf{} & \textbf{} & \textbf{\checkmark} & 0.806\textcolor{green}{$_{(+.047)}$} & 0.793 \\
(e) & \textbf{\checkmark} & \textbf{} & \textbf{\checkmark} & 0.839\textcolor{green}{$_{(+.076)}$} & 0.826 \\
CLIC & \textbf{\checkmark} & \textbf{\checkmark} & \textbf{\checkmark} & \textbf{0.893\textcolor{green}{$_{(+.134)}$}} & \textbf{0.887}
\end{tabular}}
\caption{\textbf{Ablation of CLIC.} \textbf{pre-train} denotes pre-training initialization. \textbf{IC9600} denotes fine-tuning in this dataset. \textbf{pos} denotes the positive sample selection strategy. \textbf{scm} denotes small-scale crop and merge. \textbf{CAL} is complexity-aware loss.}
\label{tab4}
\end{table}

\begin{figure}[t!]
	\centering
	\begin{minipage}{0.45\columnwidth}
		\centering
		\includegraphics[width=0.9\columnwidth]{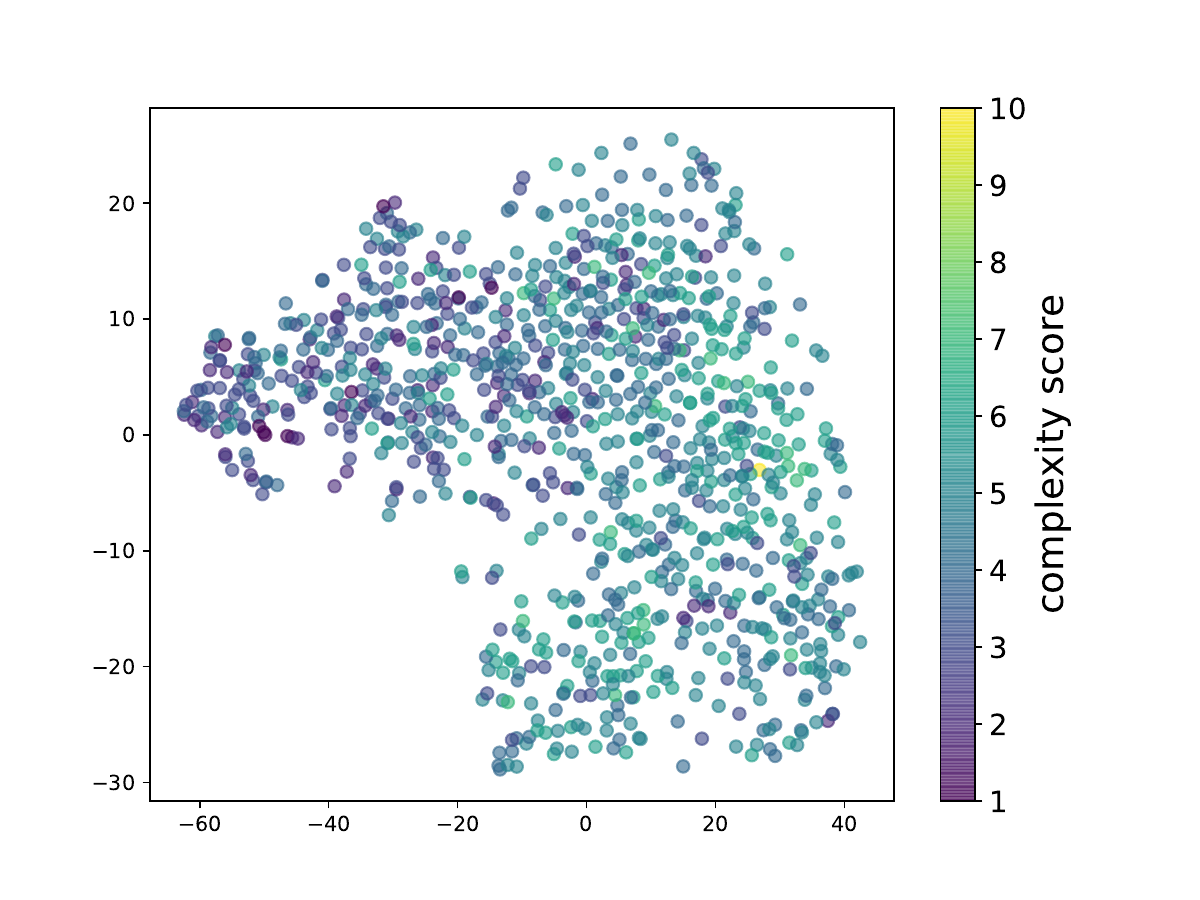}
        \subcaption{}
	\end{minipage}
	\begin{minipage}{0.45\columnwidth}
		\centering
		\includegraphics[width=0.9\columnwidth]{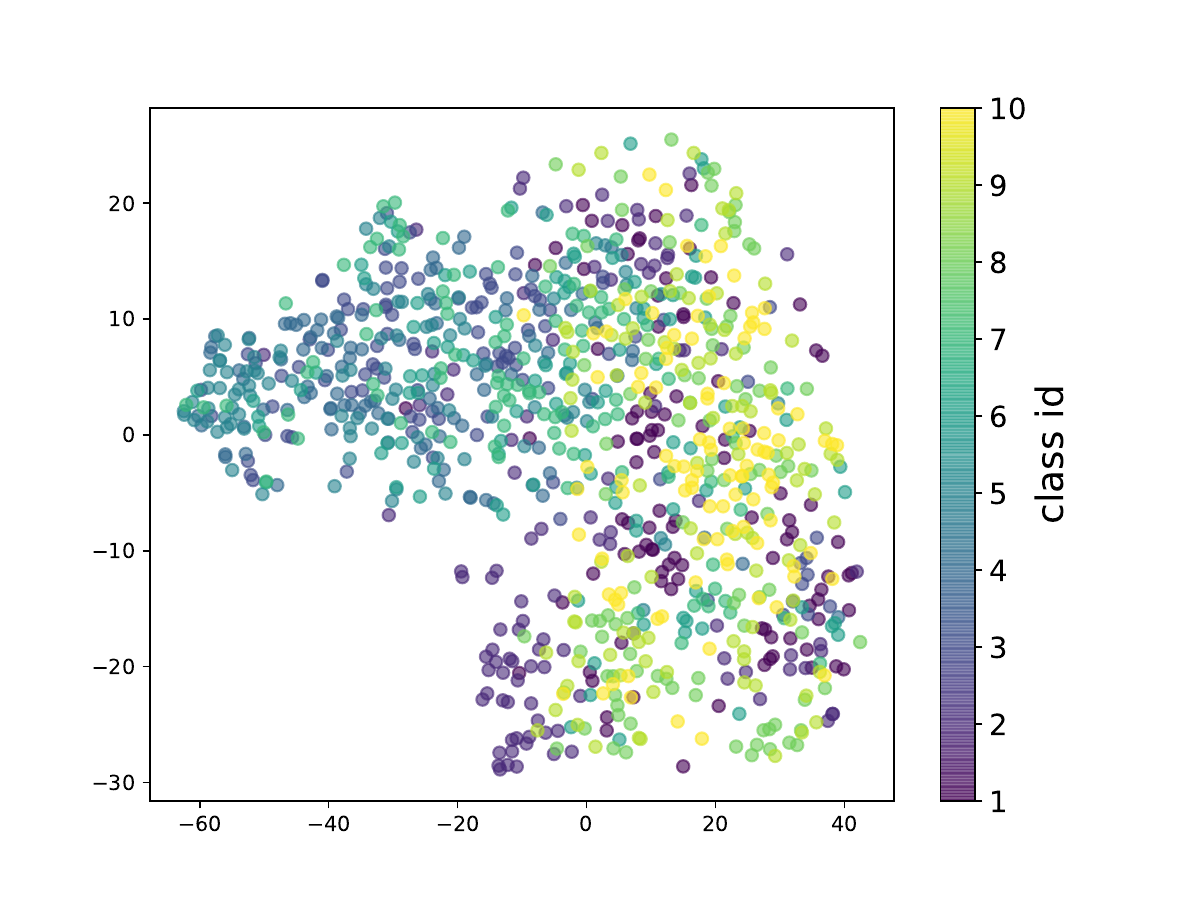}
        \subcaption{}
	\end{minipage}
\caption{\textbf{\textit{t}-SNE visualization.} All feature are taken from the output of \textbf{stage 4} in CLIC query encoder (ResNet). The \textbf{complexity score} (a) is divided into ten segments. The \textbf{class id} (b) is from 0 to 9.}
\label{fig5-tsne}
\end{figure}

We randomly sampled 1,000 untrained images from 10 classes in the ImageNet training set to perform \textit{t}-SNE visualization (Fig.\ref{fig5-tsne}). ICNet was used to infer the IC scores of these images. We then extracted the output features from stage 4 of the CLIC query encoder and reduced them to a 2-dimensional space. The images were labeled by their IC scores (divided into ten bands in the range $(0,1)$) and ImageNet class IDs, with distinct colors for each. The results show that CLIC effectively clusters samples with similar ICs, without grouping images from the same class together. This indicates that CLIC can focus on learning image complexity representations without being influenced by attributes like class. Additionally, we visualize the results of case (a) (Tab.\ref{tab4}) in Fig.\ref{fig9-case-a-tsne}.

\begin{figure}[t]
    \centering
    \includegraphics[width=0.65\columnwidth]{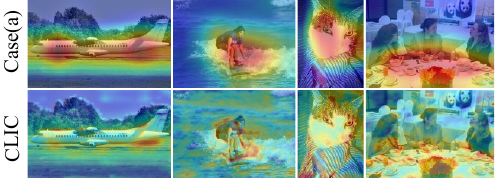}
    \caption{\textbf{Activation map.} The \textbf{case (a)} is the same in Tab.\ref{tab4}. All feature maps are extracted from stage 4 (ResNet50) of the CLIC query encoder.}
    \label{fig6-activation-map}
\end{figure}

We selected several images from the MS COCO test set and extracted the activation maps of their convolutions from the last layer of the CLIC query encoder (Fig.\ref{fig6-activation-map}). The results show that CLIC disperses hotspots across most regions of the image in the class activation map (CAM), rather than concentrating on specific objects. This demonstrates that CLIC is capable of focusing on the primary attributes of an image, rather than being biased toward individual attributes.

\subsection{Further Analysis}

\begin{figure}[t]
    \centering
    \includegraphics[width=0.65\columnwidth]{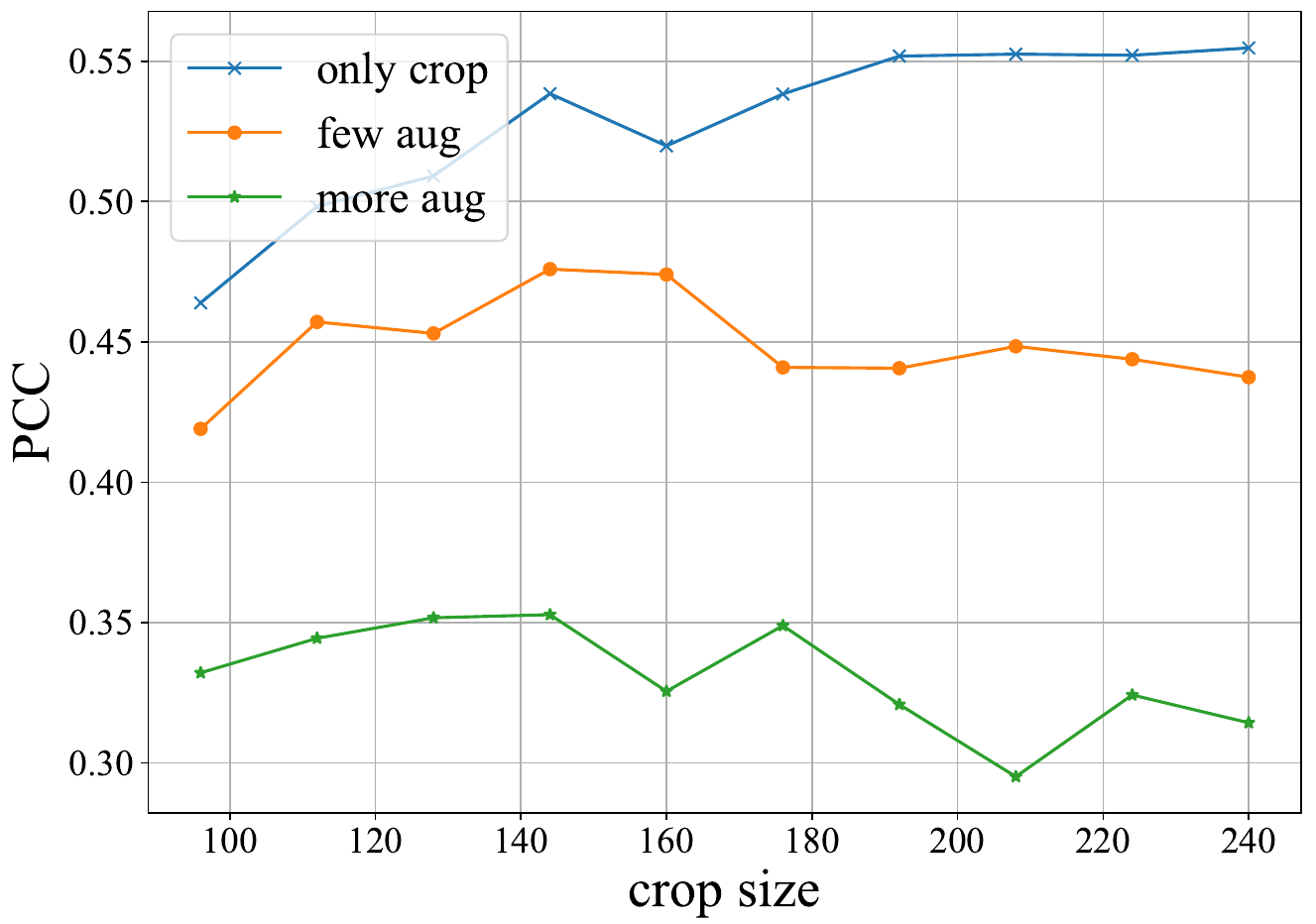}
    \caption{\textbf{Positive Views Crop Strategy.} The \textbf{only crop} denotes a crop with no data augmentation. The \textbf{few aug} denotes crops with several data augmentation. The \textbf{more aug} denotes crop with more data augmentation.}
    \label{fig7-crop-size}
\end{figure}

\textbf{Positive Views Crop Strategy.} In Sec.\ref{subsec: 3.1}, we introduced a special strategy for cropping views. For the image complexity representation task, we believe that the complexity between pairs of positive samples should be closely matched. To achieve this, we carefully designed three crop types by randomly sampling 1k images from ImageNet and processing them with crop sizes ranging from 240 to 96. We also utilized the image global entropy to evaluate these views and computed the Pearson correlation coefficient (PCC) between them and the original image entropy (Fig.\ref{fig7-crop-size}). The results indicate that all three crop strategies were appropriately sized, and when CLIC was trained based on these sizes, we obtained the expected results (Tab.\ref{tab5}). First, the augmentation methods introduced more variety, enriching the diversity of the training set, leading to continuous performance improvements. Second, we created two control sets (\textbf{case b \textit{vs.} c}, \textbf{case d \textit{vs.} e}) using suitable and unsuitable crop sizes. Although the difference between them was small, it effectively validated the usefulness of our proposed crop strategy.

\begin{figure}[t]
\centering
\includegraphics[width=0.65\columnwidth]{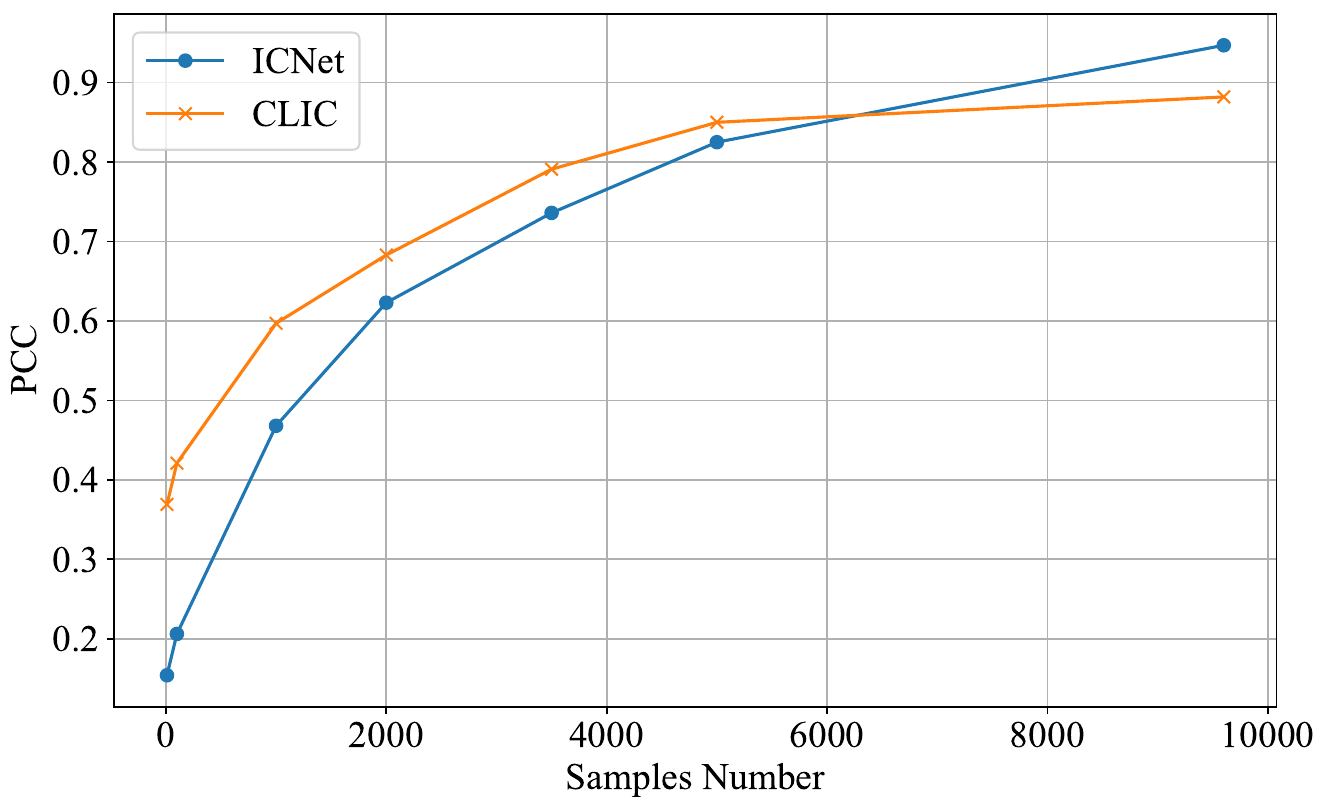}
\caption{\textbf{Performance with different labels number.}}
\label{fig8-labels-num}
\end{figure}

\textbf{Contributions of Labels.} We compared CLIC with supervised ICNet by gradually increasing the number of labels used during CLIC fine-tuning, with the results presented in Fig.\ref{fig8-labels-num}. Note that ICNet was trained from scratch with the same number of labels, rather than fine-tuned. Initially, both methods show improved performance with an increasing number of labels. Notably, CLIC achieves a PCC close to 0.4 with just ten labels, while ICNet only reaches around 0.1. As the number of labels increases, CLIC gradually approaches saturation after 5k labels, since only the header is updated during fine-tuning, while the feature extraction parameters remain fixed. However, this trend demonstrates that CLIC can rapidly improve image complexity characterization with even a small number of labeled samples, validating the effectiveness of our approach. Ultimately, we observe that when a sufficiently large number of labels is used, the supervised learning method (ICNet) significantly outperforms CLIC in accuracy.

\begin{table}[t]
\centering
\begin{tabular}{c|l|cc}
case & \multicolumn{1}{c|}{aug \& crop} & PCC↑ & SRCC↑ \\ \specialrule{1.2pt}{0pt}{0pt}
(a) & oc 224; oc 224 & 0.759 & 0.748 \\
(b) & oc 224; fa 144 & \textbf{0.771} & \textbf{0.759} \\
(c) & oc 224; fa 224 & 0.768 & 0.757 \\
(d) & fa 144; ma 144 & \textbf{0.796} & \textbf{0.781} \\
(e) & fa 144; ma 144 & 0.790 & 0.774
\end{tabular}
\caption{\textbf{Comparison of different positive view selections.} The \textbf{aug \& crop} denotes the augmentation type and crop type with the corresponding size. The \textbf{oc} is only crop, \textbf{fa} is few augmentation crop, and \textbf{ma} is more augmentation crop.}
\label{tab5}
\end{table}

\textbf{Complexity-Aware Loss.} To evaluate the role of global entropy in CAL, we compared three cases: compress ratio, edge density, and no prior information (Tab.\ref{tab6}). The results show that the performance without CAL is relatively low, with a PCC of 0.759. When edge density and global entropy were used as prior information in CAL, the PCC improved by 0.22 and 0.47, respectively, indicating that metrics with higher relevance to the ground truth are effective in enhancing performance. Conversely, the compress ratio, which has a low (or even negative) correlation with the ground truth, led to training failure. The best performance was achieved with global entropy, as the hyperparameter for edge density was not well optimized. If the edge density hyperparameter were chosen better, its performance would be closer to global entropy. Additionally, we conducted an ablation study on the $\lambda$ parameter of CAL and identified the optimal value as 0.25 (Tab.\ref{tab7}).

\begin{table}[t]
\centering
\begin{tabular}{c|ll}
Metrics & \ \ \ \ PCC↑ & \ \ \ \ SRCC↑ \\ \specialrule{1.2pt}{0pt}{0pt}
None & 0.759 & 0.748 \\
compress ratio & \ \ \ \ - & \ \ \ \ - \\
edge density & 0.781\textcolor{green}{$_{(+.022)}$} & 0.771\textcolor{green}{$_{(+.023)}$} \\
global entropy & 0.806\textcolor{green}{$_{(+.047)}$} & 0.793\textcolor{green}{$_{(+.045)}$}
\end{tabular}
\caption{\textbf{Comparison of different prior metrics.} \textbf{None} means no prior metrics.}
\label{tab6}
\end{table}

\begin{table}[t]
\centering
\begin{tabular}{c|ccccc}
$\lambda$ & 0     & 0.15  & 0.25           & 0.35  & 0.5   \\ \specialrule{1.2pt}{0pt}{0pt}
PCC↑      & 0.759 & 0.788 & \textbf{0.806} & 0.790 & 0.774
\end{tabular}
\caption{\textbf{Ablation of CAL $\lambda$}}
\label{tab7}
\end{table}

\begin{figure}[t!]
	\centering
	\begin{minipage}{0.45\columnwidth}
		\centering
		\includegraphics[width=0.9\columnwidth]{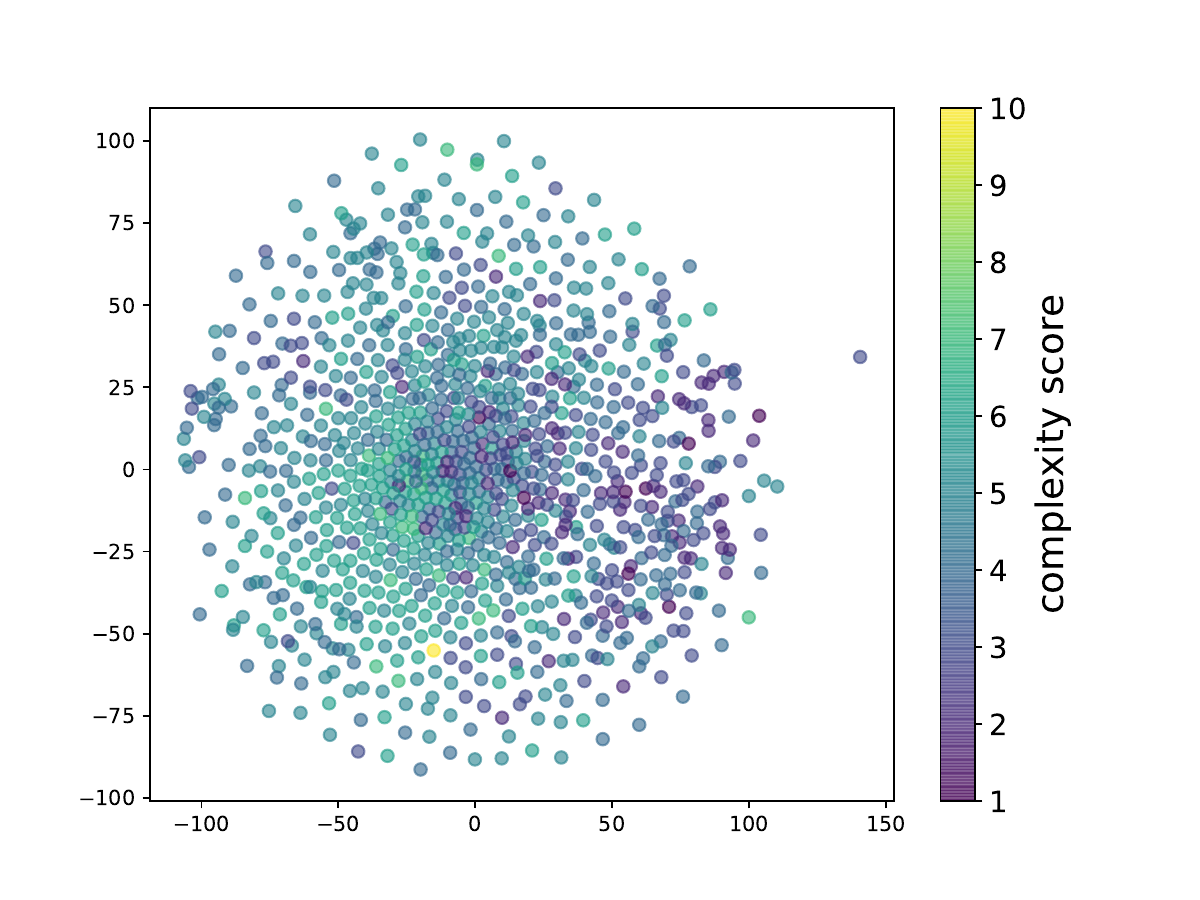}
        \subcaption{}
	\end{minipage}
	\begin{minipage}{0.45\columnwidth}
		\centering
		\includegraphics[width=0.9\columnwidth]{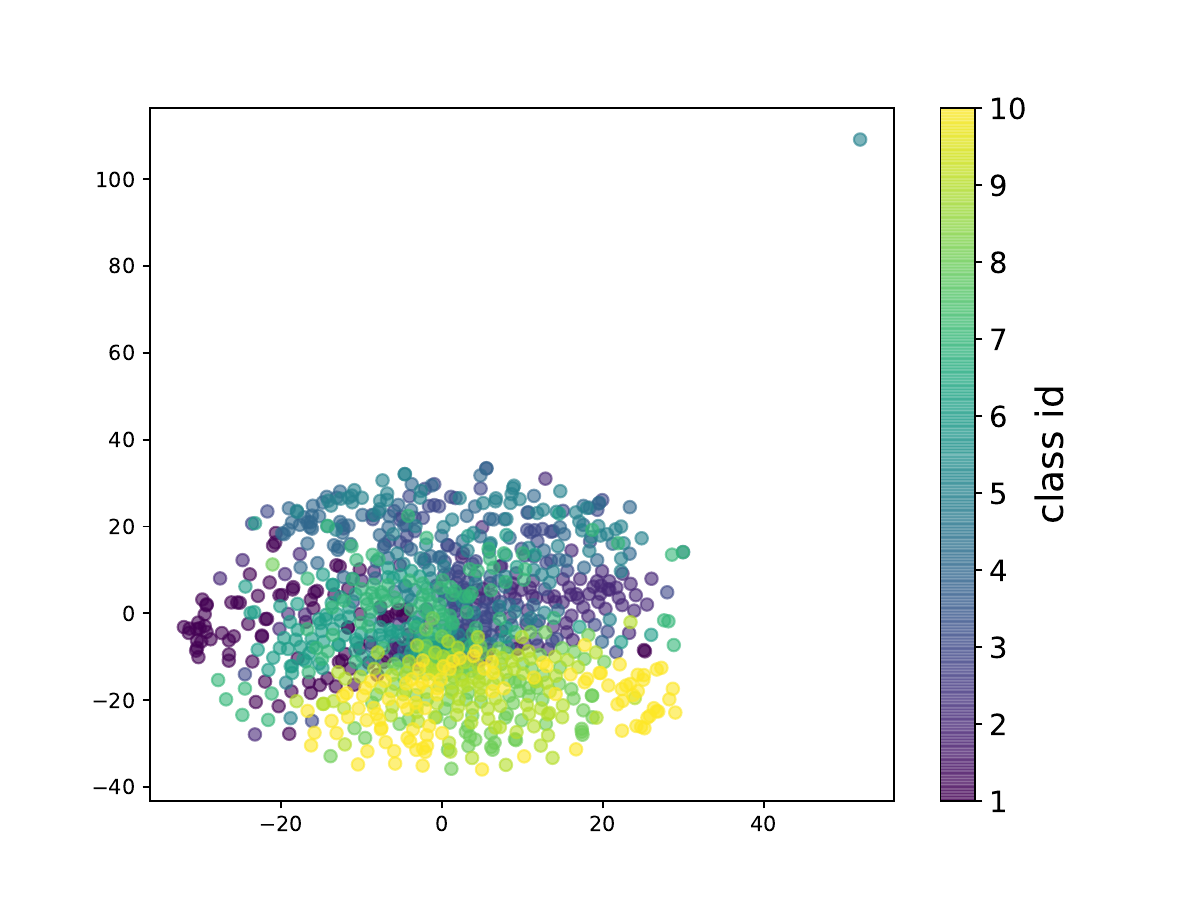}
        \subcaption{}
	\end{minipage}
\caption{\textbf{\textit{t}-SNE visualization of case (a).} All feature are taken from the output of \textbf{stage 4} in CLIC query encoder (ResNet). The \textbf{complexity score} (a) is divided into ten segments. The \textbf{class id} (b) is from 0 to 9.}
\label{fig9-case-a-tsne}
\end{figure}

\textbf{\textit{t}-SNE Visualization of Case (a).} We downscale and visualize the stage 4 output features of the CLIC query encoder, as shown in Fig.\ref{fig5-tsne}. The results indicate that CLIC effectively learns the image complexity representation without interference from image categories. For comparison, we create a control group and perform the same procedure for case (a) in Tab.\ref{tab5}, with the result presented in Fig.\ref{fig9-case-a-tsne}. It is evident that case (a) fails to learn the image complexity representation effectively when compared to CLIC.

\textbf{Decoupling Analysis of Training Data and CAL.} Global entropy was used as both a sampling basis for constructing the training set and as part of the image complexity-aware loss function. Intuitively, one might expect a strong coupling between these two components. To investigate this, we designed multiple controlled experiments to demonstrate that no such strong coupling exists and that other metrics can also yield comparable results (Tab.\ref{tab10}). Initially, we built a training set using random sampling for CLIC, and the resulting PCC and SRCC were 0.809 and 0.792, respectively. When applying CAL, the PCC and SRCC increased by 0.032 and 0.035, respectively, indicating that CAL alone significantly enhances the results. Additionally, we tested three metrics—compression ratio, edge density, and global entropy—by sampling data from both uniform and Gaussian distributions. The results showed that edge density and global entropy were similar in their performance. In contrast, the compression ratio had a negative correlation with the ground truth, leading to poorer results. We sampled using a Gaussian distribution because most IC scores in the IC9600 dataset are concentrated between 0.3 and 0.7. However, the standard Gaussian distribution primarily concentrates between 0.4 and 0.6, which resulted in fewer samples of other complexities and consequently reduced the final results compared to the uniform distribution.

\begin{table}[t]
\centering
\setlength{\tabcolsep}{1mm}{
\begin{tabular}{cl|ll}
    \multicolumn{2}{c|}{Method} & \ \ \ \ PCC↑ & \ \ \ \ SRCC↑ \\ \specialrule{1.2pt}{0pt}{0pt}
    \multirow{2}{*}{Random} & w/o CAL & 0.809 & 0.792 \\
     & w/ CAL & 0.841\textcolor{green}{$_{(+.032)}$} & 0.827\textcolor{green}{$_{(+.035)}$} \\ \hline
    \multirow{3}{*}{Uniform} & compress ratio & 0.623 & 0.604 \\
     & edge density   & 0.886 & 0.881 \\
     & global entropy & 0.893 & 0.887 \\ \hline
    \multirow{2}{*}{Gaussian} & edge density & 0.803 & 0.856 \\
     & global entropy & 0.819 & 0.812
    \end{tabular}}
    \caption{\textbf{Ablation of train data and CAL.} \textbf{w/} or \textbf{w/o} denotes that complexity aware loss is used or not. The \textbf{uniform} and \textbf{gaussian} denote uniform and gaussian distributions, respectively.}
    \label{tab10}
\end{table}

\textbf{Small-scale Crop and Merge.} The results are shown in Tab.\ref{tab11}. As the value of $c$ increases, the size of the training set expands rapidly, growing from the original 1× to 18×. We observe that the highest PCC and SRCC are achieved when $c$ is set to 3, with values of 0.906 and 0.900, respectively. However, both PCC and SRCC begin to decline significantly as $c$ increases further. This phenomenon can be explained by the fact that as $c$ grows, the training set becomes dominated by generated images, which can be considered pseudo images. The overabundance of pseudo images interferes with the contribution of real data, thereby reducing the model's performance.

\begin{table}[]
\centering
\begin{tabular}{c|c|cc}
\textit{$c$} & ratio of train set & PCC↑           & SRCC↑          \\ \specialrule{1.2pt}{0pt}{0pt}
0          & 1×                 & 0.890          & 0.883          \\
2          & 7×                 & 0.902          & 0.897          \\
3          & 11×                & \textbf{0.906} & \textbf{0.900} \\
4          & 14×                & 0.881          & 0.877          \\
5          & 18×                & 0.846          & 0.839         
\end{tabular}
\caption{\textbf{Ablation of Small-scale Crop and Merge.}}
\label{tab11}
\end{table}

\begin{figure*}[ht!]
    \centering
    \includegraphics[width=0.9\textwidth]{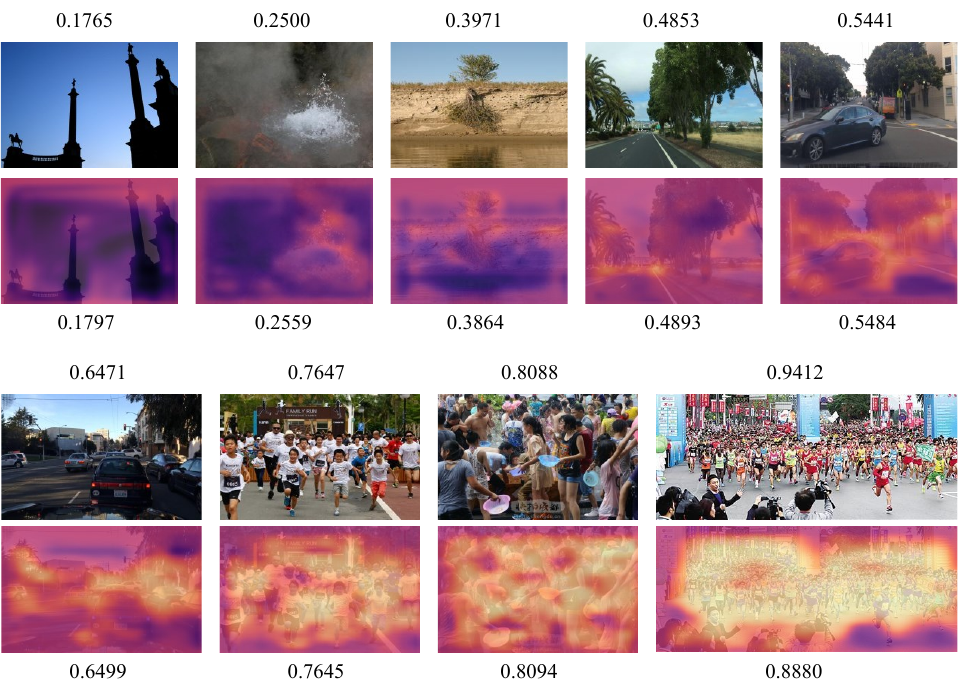}
    \caption{\textbf{Visualization of image complexity map and score.} We select several images from the IC9600 test set with image complexity scores of 0 $\sim$ 1, and use CLIC query encoder fine-tuning to get their IC map and IC score. The top image of each images pair is the original image, and the bottom image is the IC map. Similarly, the ground truth score is on top and the prediction score is on the bottom.}
    \label{fig10-ic-map}
\end{figure*}

\textbf{Image Complexity Assessment.} We select images with varying difficulty scores (Fig.\ref{fig10-ic-map}) and compute their image complexity maps and scores using the pre-trained CLIC model and ICNet after fine-tuning. The results demonstrate that our framework performs effectively in evaluating image complexity across different levels of difficulty.

\subsection{Results of Apply to Downstream Tasks.}

\begin{figure}[t!]
\centering
\includegraphics[width=0.65\columnwidth]{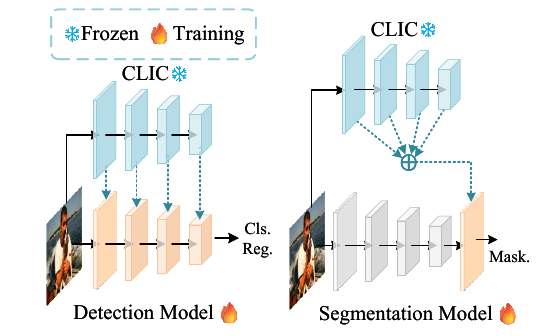}
\caption{\textbf{Visualization} of FCN with PASCAL VOC 2012 val set.}
\label{fig11-app-pipe}
\end{figure}

\textbf{Application Pipelines.} The pipelines for applying to downstream tasks are shown in Fig.\ref{fig11-app-pipe}. Mask R-CNN and FCN are used as baseline models for the detection and segmentation tasks, respectively.

\textbf{Object Detection.} We apply CLIC to Mask R-CNN \cite{maskrcnn} with several backbones. The training details follow those in MMDetection \cite{mmdet}. The results are shown in Table 10. Overall, incorporating image complexity features improves the performance of the detection model. However, as the backbone of Mask R-CNN becomes deeper, the performance improvement brought by CLIC diminishes. We hypothesize that this is due to the ResNet18-based CLIC network having fewer layers and thus being less capable of extracting image features compared to ResNet50 and ResNet101. To address this, we apply CLIC based on ResNet152 to the X-101-32x4d \cite{resnext} (Tab.\ref{tab8} X-101-32x4d$^{\dagger\dagger}$). We expect that a deeper network will better capture image complexity features. The results confirm our hypothesis, with the performance improvements for X-101-32x4d$^{\dagger\dagger}$ surpassing those of X-101-32x4d$^{\dagger}$.

\begin{table}[ht!]
\centering
\setlength{\tabcolsep}{1mm}{
\begin{tabular}{l|lll}
Backbone      & \ \ \ \ AP         & \ \ \ \ AP$_{50}$    & \ \ \ \ AP$_{75}$    \\ \specialrule{1.2pt}{0pt}{0pt}
R-50 \cite{resnet}          & 38.2       & 58.5       & 41.4       \\
R-10 \cite{resnet}          & 40.0       & 60.5       & 44.0       \\
X-101-32x4d \cite{resnext}  & 41.9       & 62.5       & 45.9       \\ \hline
R-50$^{\dagger}$ \cite{resnet}         & 38.5\textcolor{green}{$_{(+0.3)}$} & 59.1\textcolor{green}{$_{(+0.6)}$} & 41.8\textcolor{green}{$_{(+0.4)}$} \\
R-101$^{\dagger}$ \cite{resnet}        & 40.2\textcolor{green}{$_{(+0.2)}$} & 60.9\textcolor{green}{$_{(+0.4)}$} & 44.2\textcolor{green}{$_{(+0.2)}$} \\
X-101-32x4d$^{\dagger}$ \cite{resnext}  & 42.0\textcolor{green}{$_{(+0.1)}$} & 62.7\textcolor{green}{$_{(+0.2)}$} & 45.8$_{(-0.1)}$ \\
X-101-32x4d$^{\dagger\dagger}$ \cite{resnext} & 42.3\textcolor{green}{$_{(+0.4)}$} & 63.3\textcolor{green}{$_{(+0.8)}$} & 46.5\textcolor{green}{$_{(+0.6)}$}
\end{tabular}}
\caption{\textbf{Comparison of Mask R-CNN on MS COCO.} All models are trained with 1× learning rate schedule. $\dagger$ indicate CLIC is applied. The CLIC query encoder weight of feature map which is add to FPN is 0.5. CLIC is a ResNet18. $\dagger\dagger$ means CLIC is a ResNet152.}
\label{tab8}
\end{table}

\begin{table}[ht!]
\centering
\begin{tabular}{l|cll}
Method & Mem   & \ \ \ \ FPS             & \ \ \ \ mIoU         \\ \specialrule{1.2pt}{0pt}{0pt}
FCN \cite{fcn}    & 5.6G  & 19.64           & 66.53        \\
FCN$^{\dagger}$ \cite{fcn}   & 14.5G & 15.26$_{(-4.38)}$ & 68.51 \textcolor{green}{$_{(+1.98)}$} \\
FCN$^{\dagger\dagger}$ \cite{fcn}  & 14.5G & 17.83$_{(-1.81)}$ & 68.50 \textcolor{green}{$_{(+1.97)}$}
\end{tabular}
\caption{\textbf{Comparison of FCN on PASCAL VOC 2012.} The backbone of FCN is R-50-D8. $\dagger$ indicate CLIC is applied. The CLIC query encoder weight of feature map which is add to FCN head is 0.5. CLIC query encoder is a ResNet101. FCN$^{\dagger\dagger}$ indicates the concurrent processing of CLIC and FCN.}
\label{tab9}
\end{table}

\begin{figure}[t!]
\centering
\includegraphics[width=0.65\columnwidth]{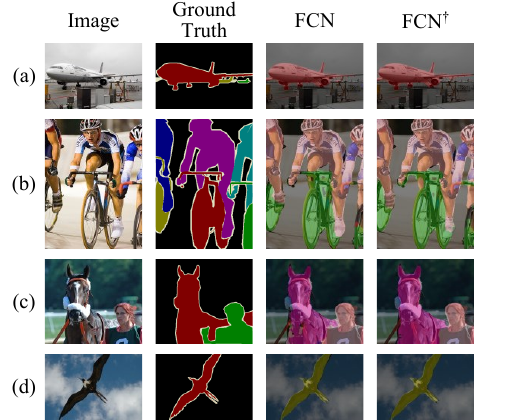}
\caption{\textbf{Visualization} of FCN with PASCAL VOC 2012 val set.}
\label{fig12-voc-vis}
\end{figure}

\textbf{Semantic Segmentation.} We apply CLIC to FCN \cite{fcn} with R-50-D8. The training details follow those in MMSegmentation \cite{mmseg}. Compared to the baseline FCN, the CLIC-assisted FCN$^{\dagger}$ shows an improvement of 1.98 mIoU, demonstrating the beneficial effect of image complexity optimization on semantic segmentation tasks. However, we observe that CLIC increases GPU memory consumption by approximately 9GB, and the FPS drops by 4.38. This represents a drawback of CLIC. Both training and inference in FCN require the use of CLIC, as the knowledge from CLIC cannot be transferred directly to FCN. While concurrent execution (FCN$^{\dagger\dagger}$) can improve the running speed, the improvement is minimal. Additionally, while GPU memory usage is similar for both concurrent and sequential execution, the CPU and regular memory must handle the extra load, which is undesirable for practical engineering applications. This is an area we plan to address in future work. We visualize some results from the PASCAL VOC 2012 validation set in Fig.\ref{fig12-voc-vis}, where the segmentation results from FCN$^{\dagger}$ show a larger and more accurate coverage of pixels for each object.

\section{Limitations}
\label{sec:limitations}
This paper proposes CLIC, an unsupervised image complexity representation learning framework based on comparative learning. Unlike previous approaches that rely on well-labeled datasets, CLIC can efficiently learn image complexity (IC) features from unlabeled data. Through extensive experimental validation, we demonstrate that CLIC can effectively learn IC representations and performs competitively with supervised learning methods when fine-tuned on the IC9600 dataset. Furthermore, CLIC significantly improves performance in several downstream tasks, showcasing its potential for real-world applications.

However, we acknowledge a few limitations of CLIC:
\begin{itemize}
    \item First, while we introduce image prior information as a supervised signal to construct an auxiliary task that aids CLIC in learning image complexity, the contrastive loss still relies on InfoNCE. This may lead CLIC to inadvertently focus on certain categories or object attributes, which could hinder the pure learning of image complexity. 
    \item Second, when applying CLIC to downstream tasks, the query encoder needs to be parallelized with the backbone models. This requirement can result in a loss of efficiency during downstream task execution.
\end{itemize}

For future work, we have two main directions. First, we aim to develop a more suitable pretext task for image complexity representation learning, which will be more tightly coupled than the traditional contrastive loss. Second, we plan to establish a framework for image complexity knowledge distillation, which would allow the IC knowledge from the query encoder to be transferred to downstream task models. This approach will maintain the efficiency of downstream tasks while enhancing their performance.

\section{Conclusion}
Image complexity assessment plays a crucial role in optimizing computer vision tasks. This paper introduces CLIC, an unsupervised image complexity representation learning framework based on comparative learning. Unlike traditional approaches that depend on well-labeled datasets, CLIC efficiently learns image complexity (IC) features from unlabeled data. Extensive experimental validation demonstrates that CLIC can effectively learn IC representations, performing competitively with supervised learning methods when fine-tuned on IC9600. Furthermore, CLIC shows substantial improvements in several downstream tasks, highlighting its potential for real-world applications. Looking ahead, our future work will focus on developing more suitable pretext tasks for IC representation learning, ensuring that these tasks remain free from interference by attributes such as categories or objects. Additionally, we aim to create an image complexity knowledge transfering framework that could serve as a "free lunch" for optimizing downstream tasks, further improving efficiency and performance.

\section{Acknowledgements}
This work would not have been possible without financial support from the National Natural Science Foundation of China (No.51209167; 52178393; 51578447), Science and Technology Innovation Team of Shaanxi Innovation Capability Support Plan (2020TD005), Xi'an Scientists \& Engineers Workforce Building Project (2024JH-KGDW-0112). The authors are also grateful to the editors and anonymous reviewers for their sound and valuable suggestions on the manuscript.

\appendix
\setcounter{section}{0}
\setcounter{figure}{0}

\section{Datasets}
\label{supp-datasets}
\subsection{ImageNet}
ImageNet \cite{imagenet} is a large-scale visual dataset with 1.28 million images in 1000 classes, widely used in computer vision and deep learning. The images' high complexity and rich visual features, including different lighting conditions, background complexity, angle diversity, etc., make it a high-quality resource for visual representation learning.

\subsection{Flickr}
Flickr \cite{flickr} contains $\sim$ 5 billion images of multiple types from Flickr (a popular online photo and video hosting, sharing, and management platform), covering a rich diversity of scenes and content. These images range from common scene types, such as natural scenes from everyday life, portraits, buildings, and city streetscapes, to more creatively expressive content, such as still life and artistic photography. The complexity of the images shows significant differences in the number of visual elements, distribution, texture, light and shadow changes, etc., which is suitable for image complexity analysis and evaluation.

\subsection{IC9600}
IC9600 \cite{ic9600} is a large-scale benchmark dataset for automated image complexity assessment. The dataset consists of 9600 carefully labeled images, each of which has been labeled in detail by 20 human annotators. The images cover a wide range of domains such as abstraction, advertising, architecture, objects, paintings, people, scenes and transportation. 

\begin{figure}[t]
\centering
\includegraphics[width=0.65\columnwidth]{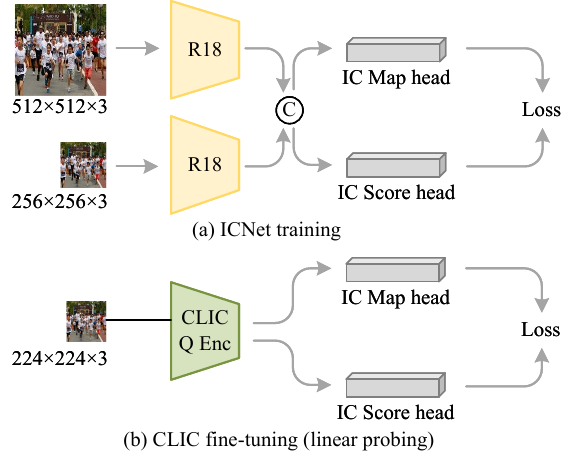}
\caption{\textbf{CLIC fint-tuning pipeline.}}
\label{fig13}
\end{figure}

\section{Metrics Details}
\label{supp-metrics}
\subsection{Unsupervised Activation Energy}
Intermediate convolutional layers in deep neural networks act as feature extractors, capable of capturing both low-level (e.g., edges, texture) and high-level features (e.g., objects and scene content) in an image. Driven by this, they define a metric, the activation energy, to quantify the visual complexity of each layer by averaging the activation values (i.e., the feature map) of the convolutional layers.

\begin{equation}
    UAE = \frac{1}{{h \times w \times d}}\sum\limits_{i,j,k}^{h \times w \times d} {F[i,j,k]}
    \label{supp-eq1}
\end{equation}

\subsection{Edge Density}

\begin{equation}
    ED = \frac{\sum_{i=1}^{H} \sum_{j=1}^{W} \mathbb{I}(\text{Canny}(I(i, j)) > 0)}{H \times W}
    \label{supp-eq2}
\end{equation}
where, $I(i,j)$ is the pixel value at position $(i,j)$ in image $I$. $\mathbb{I}(\cdot)$ is the indicator function, which returns 1 when Canny detects an edge, and 0 otherwise.

\subsection{Compress Ratio}
\begin{equation}
    CR = \frac{B}{H(I)} \quad \\ \quad H(I) = - \sum_{x=0}^{255} p(x) \log_2 p(x)
    \label{supp-eq3}
\end{equation}
where, $B$ is the bit depth of the image. $H(I)$ is the entropy of the image $I$, which can be represented by computing the probability distribution $p(x)$ of the image pixel values, $x$ is the pixel value.

\section{Fine-tuning pipeline}
The fine-tuning pipeline of CLIC for image complexity evaluation is in Fig.\ref{fig13}. ICNet contains two ResNet branch (R18) to capture detail ($512 \times 512 \times 3$) and context ($256 \times 256 \times 3$) features, respectively. Followed by IC map head and IC fraction head, which are used to calculate the loss. CLIC fine-tuning only uses the query encoder (Q Enc) as a separate branch and adopts an input size of $224 \times 224 \times 3$, followed by the structure is same as ICNet.

\bibliographystyle{elsarticle-num}
\bibliography{refs}

\end{document}